\documentclass[lettersize,journal]{IEEEtran}
\usepackage{amsmath,amsfonts}
\usepackage{algorithmic}
\usepackage{algorithm}
\usepackage{array}
\usepackage[caption=false,font=normalsize,labelfont=sf,textfont=sf]{subfig}
\usepackage{textcomp}
\usepackage{stfloats}
\usepackage{url}
\usepackage{verbatim}
\usepackage{graphicx}
\usepackage{cite}
\usepackage{xspace}

\newcommand{\etal}{\textit{et al.\xspace}}

\begin{document}

\title{VERA: Generating Visual Explanations of Two-Dimensional Embeddings via Region Annotation}

\author{Pavlin G. Poli\v{c}ar, Bla\v{z} Zupan
\thanks{P. G. Poli\v{c}ar is with the Faculty of Computer and Information Science, University of Ljubljana.}
\thanks{B. Zupan is with the Faculty of Computer and Information Science, University of Ljubljana and the Baylor College of Medicine.}}

\markboth{}%
{Poli\v{c}ar and Zupan: VERA: Generating Visual Explanations of Two-Dimensional Embeddings via Region Annotation}

\IEEEpubid{}

\maketitle

\begin{abstract}
Two-dimensional embeddings obtained from dimensionality reduction techniques such as MDS, t-SNE, or UMAP, are widely used to visualize high-dimensional data and support researchers in visually identifying clusters, outliers, and other interesting patterns in the data. However, the main challenge is not only to detect such patterns, but to explain what they represent in terms of the original, human-interpretable features of the data. Existing approaches often rely on interactive exploration or direct feature encodings, requiring substantial manual inspection that can be time-consuming and repetitive.
As an alternative, we propose VERA (Visual Explanations via Region Annotation), a general-purpose method for explaining two-dimensional embeddings through automatically generated, static, region-based visual explanations.
VERA identifies informative regions in the embedding space and associates them with user-provided human-interpretable features, producing concise visual annotations that summarize the structure of the embedding landscape at a glance. Rather than merely showing where feature values occur, VERA automatically filters, merges, and ranks candidate explanations, enabling users to focus on the most informative embedding structures without manual exploration.
We demonstrate VERA's utility on several real-world datasets and evaluate its effectiveness in a user study comparing it with the utility of a comprehensive interactive data mining toolkit.
Our results show that VERA's generated static explanations can convey the essential insights of complex embeddings and support users in typical exploratory data analysis tasks, while requiring significantly less time and user effort.
\end{abstract}


\section{Introduction}
\begin{figure}[htb]
  \centering
  \includegraphics{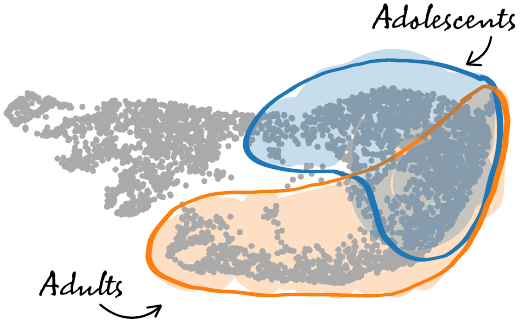}
  \caption{\label{fig:motivation}
An example of a two-dimensional embedding and its possible interpretation. The UMAP visualization shows the relationships between 4,177 marine mollusks. After obtaining the embedding, the next step is to determine what the shapes and clusters represent. For example, we might determine that the mollusks in the blue region predominantly correspond to adolescent animals while those in the orange region correspond to mature mollusks.
}
\end{figure}

\IEEEPARstart{T}{wo-dimensional}, point-based visualizations of high-dimensional data are a cornerstone of exploratory data analysis and are used across diverse scientific domains such as biology, economics, chemistry, political science, astronomy, and physics~\cite{Liu2017,DeBodt2025}. As multivariate datasets with possibly thousands of features and a large number of samples have become commonplace, visual exploration is often the first step toward discovering meaningful patterns and generating hypotheses. In practice, however, interpretation workflows typically focus on subsets of features that are human-interpretable and relevant to the analysis task, often ranging from tens to a few hundred variables. Our work is designed for this setting, where analysts seek concise explanations in terms of a manageable number of features rather than exhaustively processing thousands of dimensions. Visualizations of such datasets, often called \textit{embeddings}, project high-dimensional relationships into a two-dimensional space, as illustrated in Fig.~\ref{fig:motivation}. Once the embedding is obtained, analysts seek to interpret what its shapes and clusters represent in the original data.

Today, data science practitioners have at their disposal a plethora of scalable dimensionality reduction methods that can be used to reduce data to two dimensions, including principal components analysis (PCA)~\cite{Jolliffe2002}, linear discriminant analysis (LDA)~\cite{Fisher1936}, FreeViz~\cite{Demsar2007}, multidimensional scaling (MDS)~\cite{Kruskal1978}, t-distributed stochastic neighbor embedding (t-SNE)~\cite{Maaten2008}, and uniform manifold approximation and projection (UMAP)~\cite{McInnes2018}.
These methods are routinely applied to reveal clusters, transitions, and outliers that provide insight into the underlying data's landscape.

After performing dimensionality reduction, analysts typically examine the resulting visualization to identify any discernible visual patterns and interpret how these reflect underlying phenomena in the data.
Linear methods, such as PCA, LDA, and FreeViz, offer intuitive interpretations since their inferred dimensions correspond to linear combinations of the original features. Nonlinear methods, such as MDS, t-SNE, and UMAP, can reveal more complex patterns and relationships, yet the lack of direct correspondence between embedding dimensions and original features makes interpretation difficult. As scalable dimensionality reduction algorithms continue to improve~\cite{McInnes2018,Linderman2019,Lambert2022}, analysts frequently generate multiple embeddings using different parameter settings or algorithms, each revealing distinct aspects of the underlying data manifold~\cite{Kobak2019,DeBodt2025}. Each resulting embedding must then be inspected and interpreted, a process that is time-consuming, subjective, and cognitively demanding.

Existing research on interpreting two-dimensional embeddings has focused primarily on interactive visualization tools that allow users to explore the embedding space in detail. Examples include systems that link embeddings to feature distributions, enable brushing and linking, or provide cluster exploration interfaces (e.g.,~\cite{Endert2014,Krause2014}).
While these tools enable an in-depth analysis, the main goal of exploratory data analysis (EDA) is often to obtain a concise, high-level overview of the embedding structure rather than to investigate every local pattern interactively.

Although many tools offer features to speed up and assist in the exploration process, e.g., automatic clustering or feature ranking, analysts must still actively navigate and make sense of the embedding, activities that are both repetitive and time-intensive.
A typical workflow may involve selecting a group of data points, inspecting their feature distributions, comparing these to the remaining data points, and annotating the visualization accordingly. This must, of course, be repeated for each interesting group of data points in each resulting embedding.

Given the repetitive nature of these exploratory data analysis tasks, automated, static approaches that require no user interaction in explaining the data landscapes can substantially reduce analysis time and cognitive effort while still conveying the essential structure of the embedding. Such methods can complement interactive tools by providing quick, interpretable summaries that guide deeper investigation when needed.

In this paper, we propose \textit{Visual Explanations via Region Annotation (VERA)}, an automated, general-purpose method for generating static visual explanations of two-dimensional embeddings. VERA automatically identifies informative regions in the embedding space and associates them with a smaller set of user-provided human-interpretable features, presenting these associations as concise, small-multiple visual summaries. This approach helps analysts quickly and easily obtain a high-level overview of the embedding.

Our main contributions are as follows:
\begin{enumerate}
  \item We introduce VERA, a general-purpose framework for generating static, region-based visual explanations of two-dimensional data embeddings.
  \item We present an automated procedure for detecting informative regions of point-based visualizations and linking them to relevant, user-provided human-interpretable features.
  \item We validate VERA through a user study comparing it with the utility of a comprehensive interactive data mining toolkit, showing that VERA enables users to perform typical embedding interpretation tasks with comparable accuracy while requiring significantly less time.
\end{enumerate}

Together, these contributions demonstrate that VERA’s static visual explanations can effectively convey the essential insights of complex embeddings, offering a fast and accessible complement to interactive data exploration.

\section{Related Work}
\label{sec:related_work}

Alongside the development of novel dimensionality reduction approaches for the visualization of high-dimensional data~\cite{Kruskal1978,Maaten2008,McInnes2018}, several methods have been developed to aid in interpreting the resulting visualizations~\cite{ChengMueller2016,Marcilio2021,Fujiwara2020}.
General-purpose interpretation approaches aim to relate the obtained embedding to human-interpretable input features.
In domains where input features are not directly interpretable -- such as images, compositional data, or deep latent representations -- domain-specific solutions are often required, as in text~\cite{Heimerl2016,WangHohmanChau2023}, images~\cite{Liu2019,Ye2024}, and drug-prescription data~\cite{Policar2023}.

Existing techniques differ in how much of the interpretive work they automate.
Some techniques primarily encode feature values directly in the embedding, for example through color- and glyph-based encodings, spatial partitions such as Voronoi-based approaches~\cite{Broeksema2013}, and region overlays such as rangesets~\cite{Sohns2022}. These techniques show where feature values occur in the embedding, but largely leave it to the analyst to determine which patterns are most relevant to the observed structure.
Other techniques go a step further and algorithmically construct explanations by selecting, filtering, aggregating, or ranking candidate regions or feature-value patterns.
The distinction therefore lies in the primary role of the method: direct feature-value encodings expose structure for inspection, whereas explanation-generation methods additionally prioritize candidate explanations.

Among explanation-generation techniques, existing work can be broadly divided into global and local approaches.
Global approaches typically express the embedding axes as linear combinations of the original features~\cite{Broeksema2013,Coimbra2016}.
While effective for linear projections, most modern dimensionality reduction techniques rely on highly nonlinear transformations of the original feature space, making global approaches unsuitable for interpreting these embeddings.
Local approaches instead explain local patterns in the embedding, such as clusters, transitions, and outliers, and relate them to particular combinations of high-dimensional feature values.
These approaches often include explanatory overlays directly on the embedding~\cite{Joia2015,ChengMueller2016,Tian2021}, or provide interactive linked-view systems that allow users to inspect candidate explanations in more detail~\cite{Kandogan2012,Stahnke2016,Bibal2021}.
Broadly speaking, local approaches can be categorized into two types: descriptive approaches and contrastive approaches.

Descriptive approaches identify regions of the embedding space, most commonly clusters, and determine which combinations of high-dimensional feature values are characteristic of the samples contained within each region.
For instance, Joia \etal~\cite{Joia2015} identify informative features by examining the SVD loadings of each visual cluster, while Da Silva \etal~\cite{DaSilva2015} and subsequent extensions by Van Driel \etal~\cite{VanDriel2020} and Tian \etal~\cite{Tian2021} use local variance to detect features with similar values within embedding regions.
A similar interactive, variance-based approach was proposed by Pagliosa \etal~\cite{Pagliosa2016}.
Alternatively, Broeksema \etal~\cite{Broeksema2013} use Multiple Correspondence Analysis to jointly embed samples and features into a shared embedding space and identify characteristic features based on their proximity to samples. Cheng and Mueller~\cite{ChengMueller2016} apply a similar strategy based on MDS.

While descriptive approaches explain \textit{what} particular regions correspond to, contrastive approaches explain \textit{how} regions differ from one another. In a recent survey of sociological literature, Miller~\cite{Miller2019} highlights that humans prefer contrastive explanations and concludes that explanations in artificial intelligence should account for this preference.
Contrastive approaches therefore aim to identify combinations of feature values that distinguish one region from another.
Several approaches combine well-established statistical tools -- such as feature distribution plots~\cite{Stahnke2016,Eckelt2022}, statistical tests~\cite{Marcilio2021}, decision trees~\cite{Bibal2021}, and Shapley values~\cite{MarcilioEler2021} -- with intuitive, interactive user interfaces.
Others propose custom solutions; for example, Fujiwara \etal~\cite{Fujiwara2020} propose a contrastive variant of PCA, while Kandogan~\cite{Kandogan2012} combines several hand-crafted metrics.

Despite the plethora of existing approaches, current solutions share several notable limitations. Most are implemented as interactive tools that require extensive user interaction. This is especially time-consuming when inspecting and comparing multiple embeddings.
Other approaches are limited to producing a single explanation, often focusing on the most salient combination of feature values while neglecting other potentially interesting but less prominent combinations~\cite{DaSilva2015,VanDriel2020,Tian2021}, or generate a large number of visualizations that users must manually inspect~\cite{Faust2019,Plumb2020}.
Others are tightly coupled to specific dimensionality reduction techniques, reducing their general applicability across methods~\cite{Chatzimparmpas2020,Broeksema2013,ChengMueller2016}. Finally, the majority of existing approaches define regions based on a single, discrete clustering of the embedding space. This approach risks potentially missing overlapping or continuous structures in the embedding space, such as transitions or trajectories, which do not fit neatly within discrete cluster boundaries.

In contrast to existing explanation-generation methods, VERA automatically produces concise, static visual explanations without requiring user interaction. It combines descriptive and contrastive perspectives to summarize the embedding structure, providing informative regional annotations rather than a single global summary or a large set of candidate views. Unlike most existing methods, VERA does not rely on predefined clusters and can detect and annotate arbitrary visual structures in the embedding space.

\section{Methods}
\label{sec:methods}

\begin{figure*}[t]
  \centering
  \includegraphics{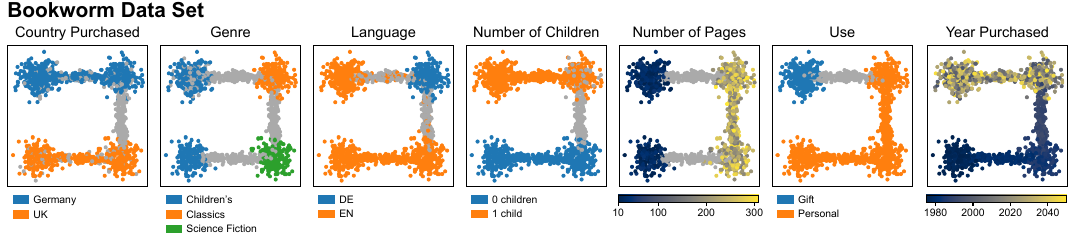}
  \caption{\label{fig:variable_values}Scatter plots for each feature of our fictional Bookworm dataset. The point positions correspond to its two-dimensional embedding, revealing four distinct clusters connected by three transitions. Categorical variables are colored with a discrete colormap, while the two numeric variables are colored with a continuous colormap. Missing values are colored gray.}
\end{figure*}

Miller~\cite{Miller2019} posits that the process of explanation involves three cognitive stages: generation, selection, and evaluation. In the context of explaining two-dimensional embeddings, VERA aims to automate the first two of these, generating several candidate explanations, ranking them by informativeness, and presenting the selected explanations for human evaluation, interpretation, and contextualization.

In line with existing work, we adopt two complementary types of visual explanations: \textit{contrastive explanations}, showing how the values of a high-dimensional feature vary across the embedding space, and \textit{descriptive explanations}, summarizing feature values characteristic of individual highlighted regions.
Contrastive explanations answer questions like ``Where do the different feature values occur?'' while descriptive explanations answer ``What feature values are characteristic for this region?''.

Given a dataset $D \in \mathbb{R}^{N \times |F|}$ and its corresponding two-dimensional embedding $Y \in \mathbb{R}^{N \times 2}$, our aim is to relate different regions of $Y$ back to the original, high-dimensional features $f \in F$.
Some feature values may be characteristic of specific regions of $Y$; for instance, data points in a particular cluster may all share the same category or contain similar numeric values.
We group informative feature values within \textit{rules} that indicate a particular category or numeric interval.
A rule $\varphi_i$ is defined as a function $\varphi: Y \rightarrow \left \{ 0, 1
\right \} $, which evaluates to $1$ if a sample $y$ satisfies the rule condition and $0$ otherwise.
Rules are associated with \textit{regions} via \textit{region annotations} $e_i$, which represent pairs of rules $\varphi_i$ and regions $r_i$, $e_i = \left ( r_i, \varphi_i \right )$.
Each region $R_i$ represents a set of points $R_i = \{ y_1, y_2, \cdots, y_N \} \subseteq Y$ belonging to that region.
Region annotations $\{ e_1, e_2, \cdots, e_k \}$ can be merged into \textit{region annotation groups} $e_{\text{group}}$ associated with rule $\varphi_{\text{group}}(y) = \prod_{i=1}^k \varphi_i(y)$ and region $R_{\text{group}} = \bigcup_{i=1}^{k} R_i$. Since region annotation groups serve the same conceptual purpose as region annotations, we use the two terms interchangeably.
Once we have identified a set of informative region annotations $E = \left \{ e_1, e_2, \cdots, e_K \right \}$, the next step is to display these to the user. Displaying all region annotations simultaneously in a single visualization would lead to overlapping regions and labels, reducing readability.
Instead, we group region annotations into \textit{panels} $P \subseteq E$ (also referred to as \textit{visual explanations}) and arrange collections of panels into \textit{layouts} $L = \left \{ P_1, P_2, \cdots, P_L \right \}$. A layout is then displayed in a small multiple fashion.

In the upcoming sections, we use a fictional, guiding example to outline the individual steps of our approach.
This fictional dataset, aptly named the Bookworm dataset, comprises samples corresponding to books owned by an imaginary literary enthusiast. Each book is characterized by seven features: language, genre, number of pages, year and country of purchase, whether the owner had children at the time of purchase, and whether the book was purchased for personal use or as a gift.
Fig.~\ref{fig:variable_values} shows their distribution in a specially designed two-dimensional embedding.

Through these books, we can trace the life of this fictional book collector. The bottom left cluster -- filled with children's books presumably given to them by their parents -- characterizes their childhood. Their teenage years are represented by the bottom right cluster, filled with longer science fiction books. After spending their formative years in the United Kingdom, the collector relocates to Germany, where their literary tastes shift to German classics, which remain their favorites to this day. Soon thereafter, they have their first child, to whom they resolve to instill an appreciation for the literary arts. However, to preserve their child's English heritage, they limit their offspring to a strict diet of English literature.
While this narrative can be inferred by carefully examining each feature shown in Fig.~\ref{fig:variable_values}, the following sections show how VERA uncovers this same narrative automatically.

\subsection{Region Construction}
\label{sec:region-construction}

\begin{figure*}[t]
  \centering
  \includegraphics{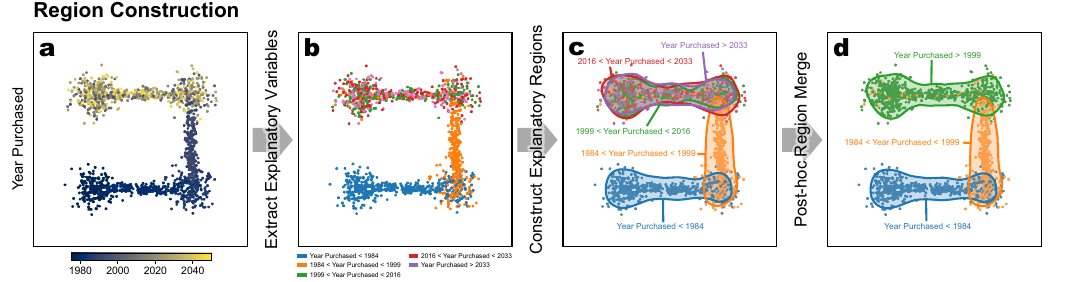}
  \caption{\label{fig:region_construction}We illustrate the region construction process for a single feature using a synthetic example. (a) Two-dimensional embeddings are often depicted by coloring points according to their feature values. (b) We discretize the numeric values into five bins, resulting in five binary indicator variables. Here, points are colored according to their discretization bin membership. (c) For each of the discretized bins, we extract regions by tracing the contours of the corresponding KDE. (d) Lastly, we apply our post-hoc merging procedure to consolidate overlapping and semantically compatible region annotations.}
\end{figure*}

Region annotations are the fundamental building blocks of visual explanations, representing groups of data points in the embedding that exhibit shared characteristics.
In this section, we outline the steps to identify and construct informative region annotations. We then discuss how to combine these regions into informative visual explanations in the next two sections.
At a high level, our region construction procedure consists of three steps, illustrated in Fig.~\ref{fig:region_construction}.

\subsubsection{Extracting Explanatory Variables}
\label{sec:discretization}

An input dataset $D$ may contain a mix of categorical and numeric features.
For categorical features, we apply one-hot encoding, representing each sample's membership in a given category using $k$ binary indicator variables.
For numeric features, we discretize their values using univariate $k$-means binning, representing each bin by a binary indicator variable that denotes whether a sample falls within the corresponding interval\footnote{We selected $k$-means binning because it better captures skewed and multimodal feature distributions, producing intervals that more closely follow the natural value groupings present in the data.}.
For example, the numeric feature in Fig.~\ref{fig:region_construction}.a is discretized into five bins shown in Fig.~\ref{fig:region_construction}.b.
We associate each resulting indicator variable with a rule $\varphi_i$ that specifies the corresponding discretization interval or category membership.

\subsubsection{Constructing Region Annotations}
\label{sec:region_construction}

Next, we locate the samples corresponding to each indicator variable within the embedding.
For each variable, we obtain its kernel density estimate (KDE), denoted by $D$, using a Gaussian kernel function.
When sample-wise distortion measures are available, we downweight poorly mapped samples during KDE estimation, making spurious regions in distorted parts of the embedding less likely to be extracted\footnote{These regions likely correspond to spurious artifacts of the dimensionality reduction process rather than any true underlying structure.}.
We extract contour lines of the KDE at a user-specified level, defaulting to $0.25 \times \max D$, and use the resulting polygons as the visual representation of a region $r_i$.
This value provides a practical middle ground: lower levels produce broader, less localized regions, whereas higher levels restrict contours to dense cores and can fragment them.
In the resulting visualizations, these contours serve as the primary representation of region annotations, while the scatter plot provides context for the underlying samples.
By construction, each resulting region encompasses samples that predominantly share a particular feature value.
We combine the obtained rules and regions into region annotations $e_i = \left( r_i, \varphi_i \right)$, which serve as the building blocks for constructing visual explanations.
In our running example, the five categories corresponding to discretization bins in Fig.~\ref{fig:region_construction}.b are used to construct the five corresponding regions shown in Fig.~\ref{fig:region_construction}.c.

An important consideration of the KDE step is the choice of Gaussian kernel bandwidth $\sigma$.
Different embedding algorithms produce embeddings that can span across different scales, e.g., one embedding may span from $-1$ to $1$ along its dimensions, while another may span from $-10$ to $10$.
Additionally, a single scale parameter is unlikely to be appropriate for datasets containing different numbers of samples. For instance, in a dataset with 50 samples, we are likely to be interested in groups of 10-20 points, while in a dataset containing 2 million samples, the groups of interest are likely larger.
By default, we set $\sigma$ to equal the median distance of each point's $k$-th nearest neighbor, setting $k=\sqrt{N}$. This yields a scale-adaptive local bandwidth that grows with the sample size while remaining sensitive to local structure.
Intuitively, the scale parameter controls the granularity of the resulting region annotations: smaller values produce many fine-grained regions, whereas larger values yield broader, higher-level regions.

Using KDE contours as regions provides several advantages: most importantly, it removes the need for a separate clustering step in the 2D embedding, which allows VERA to uncover continuous transitions rather than only cluster assignments. KDEs can also be computed efficiently, scaling linearly with the number of samples, and allow distortion measures to be incorporated directly during region construction.
We note, however, that the region construction step is modular: KDE contours can be replaced by alternative region representations while leaving the remaining steps unchanged, such as the alpha hulls used in rangesets~\cite{Sohns2022}. We include several visualizations based on this variant in the supplementary material.

\subsubsection{Post-hoc Region Merging}
\label{sec:post-hoc-merge}

Samples belonging to different intervals or categories often occupy the same regions of the embedding space, producing overlapping region annotations.
In some cases, the rules associated with these overlapping regions can be merged into a single rule.
For example, the three overlapping regions in the top part of Fig.~\ref{fig:region_construction}.c have semantically compatible rules and can be merged into a single region annotation, shown in Fig.~\ref{fig:region_construction}.d.
To identify and merge overlapping region annotations and their associated semantically compatible rules, we introduce a simple post-hoc merging procedure.
Two or more region annotations are eligible for merging if they exhibit high spatial overlap and semantic compatibility between their rules.

Determining the semantic compatibility of rules is straightforward. For rules derived from ordinal or numeric features, only adjacent bins or categories may be merged. For rules originating from nominal categorical features, any combination of categories can be merged.

Determining the degree of spatial overlap between two regions is also straightforward.
We define the \textit{maximum overlap} between two regions $r_i$ and $r_j$ as
\begin{equation}
\text{max\_overlap}(r_i, r_j) = \max \left \{ \frac{|r_i \cap r_j|}{|r_i|}, \frac{|r_j \cap r_i|}{|r_j|} \right \}, \notag
\end{equation}
where $|r|$ denotes the number of samples within that region.
A pair of regions is merged when their maximum overlap exceeds a user-specified threshold. We use a conservative default so that clearly redundant regions are consolidated while nearby but distinct structures remain separate.
This formulation ensures that small regions, which are completely encompassed by a larger region, can be absorbed into the larger region.

Following the merging step, features with little or no spatial localization usually produce a single region annotation spanning the entire embedding.
Because such regions reflect uniformly distributed feature values and carry no structural information, we discard them from further analysis.

\subsection{Contrastive Explanations}
\label{sec:contrastive}

Contrastive explanations show how the values of a particular feature vary across the embedding space. While conceptually similar to coloring points by feature value, our approach automatically filters out spatially uninformative features and ranks the remaining ones by informativeness. This allows users to focus on the most relevant features without manually sifting through a potentially large set of candidates.

\subsubsection{Explanation Construction}

We generate contrastive explanations by including all region annotations $e_i$ associated with a given feature $f$.
However, even after filtering singleton region annotations, a large number of candidate explanations may still remain.
To further reduce the number of candidates, we introduce the \textit{contrastive merge}, which combines features whose regions exhibit strong spatial correlation.

\begin{figure}[t]
  \centering
  \includegraphics{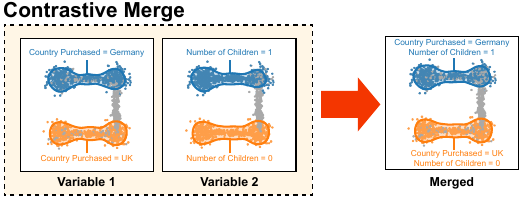}
  \caption{\label{fig:contrastive_merge}The contrastive merge. If two features contain region annotations that have a nearly perfect overlap bipartite matching, we can merge the two features into one. This not only reduces the number of panels a user needs to inspect but also reveals relationships between features, avoiding the need to compare multiple nearly identical visualizations manually.}
\end{figure}

Unlike the post-hoc merging procedure described in Section~\ref{sec:post-hoc-merge}, which merges only region annotations derived from a single feature, the contrastive merge combines region annotations derived from multiple features. To merge region annotations associated with two features $f_1$ and $f_2$, each region annotation $e_i \in f_1$ must have a nearly perfectly overlapping counterpart $e_j \in f_2$. When such a bijection exists, the matching pairs of region annotations are merged, effectively representing both features using a single collection of region annotations.
The same logic naturally extends to more than two features.
For example, the left panel in Fig.~\ref{fig:contrastive_merge} depicts two separate features whose region annotations exhibit near-perfect overlap and can be merged into a single explanation shown on the right.
This reduces the number of candidate visual explanations while revealing relationships among high-dimensional features, and avoids the need for users to mentally align nearly identical spatial patterns across panels. Instead, similar features are summarized in a single explanation, making both the redundancy and the underlying feature relationships explicit.

\subsubsection{Explanation Selection}
\label{sec:contrastive-selection}

To identify the most informative contrastive explanations, we rank and select visual explanations using three criteria: total region overlap, mean purity, and a human attention score.

Visual explanations describing features whose values are localized to particular regions of the embedding space are likely to be more informative than features whose values are spread more evenly across the embedding space. Region annotations associated with highly localized features exhibit less overlap than those of more evenly distributed features.

We compute the overlap between two region annotations as the Jaccard similarity of their regions and define the \textit{total region overlap} of a panel $P$ by taking the mean overlap between all pairs of region annotations $(e_i, e_j)$.

The \textit{purity} of each region annotation indicates what proportion of the samples encompassed by the region satisfy its associated rule:
\begin{equation}
\text{purity}(r) = \frac{1}{|r|} \sum_{y_i \in r} \varphi_i (y_i). \notag
\end{equation}
High-purity regions indicate stronger spatial structure and are therefore preferred. For each panel, we compute the \textit{mean purity} of the contained explanatory variables.

A well-known finding from psychology states that human working memory has limited capacity and can pay attention to only a handful of items at a time. Several bounds have been proposed, including seven~\cite{Miller1956}, four~\cite{Cowan2001}, and two~\cite{Gobet2004}. Panels containing many region annotations are more likely to overwhelm users; we therefore include a \textit{human attention score} as a heuristic readability prior. We define this score as $\left|3 - |P|\right|$, where $|P|$ denotes the number of region annotations in panel $P$, assigning the lowest score to panels containing three annotations. Here, three serves as a practical middle ground rather than a universal optimum.

\begin{figure}[t]
  \centering
  \includegraphics{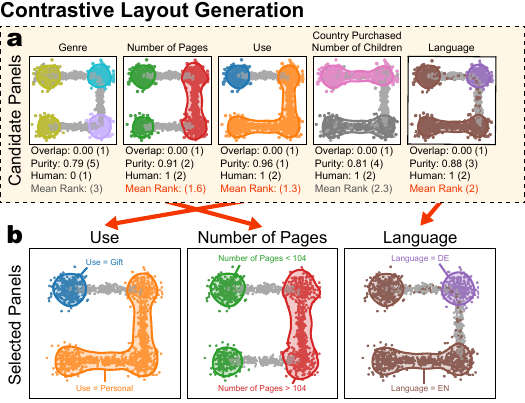}
  \caption{\label{fig:contrastive_generation}Contrastive layout generation. (a) Candidate panels are generated by including all region annotations associated with a particular feature. We score and rank each candidate panel according to three metrics and obtain a final, weighted mean rank. (b) The top $k$ panels, indicated in red, are included in the resulting layout.}
\end{figure}

We compute individual panel rankings for each of the three metrics. Because no single metric reliably produced good rankings across datasets, we aggregate them using a weighted mean-rank strategy. This treats the criteria as complementary heuristics and avoids relying on any one metric alone. We then select the top $k$ panels with the highest overall ranking as our final contrastive explanations. This process is illustrated in Fig.~\ref{fig:contrastive_generation}.

\subsection{Descriptive Explanations}
\label{sec:descriptive-explanations}

Descriptive explanations characterize regions of the embedding space by identifying combinations of features and feature values that are characteristic of the samples within those regions. Unlike contrastive explanations, which indicate where different feature values occur across the embedding space, descriptive explanations should be interpreted as local summaries of highlighted regions.

\subsubsection{Explanation Construction}
\label{sec:descriptive-explanations:construction}

\begin{figure}[t!]
  \centering
  \includegraphics{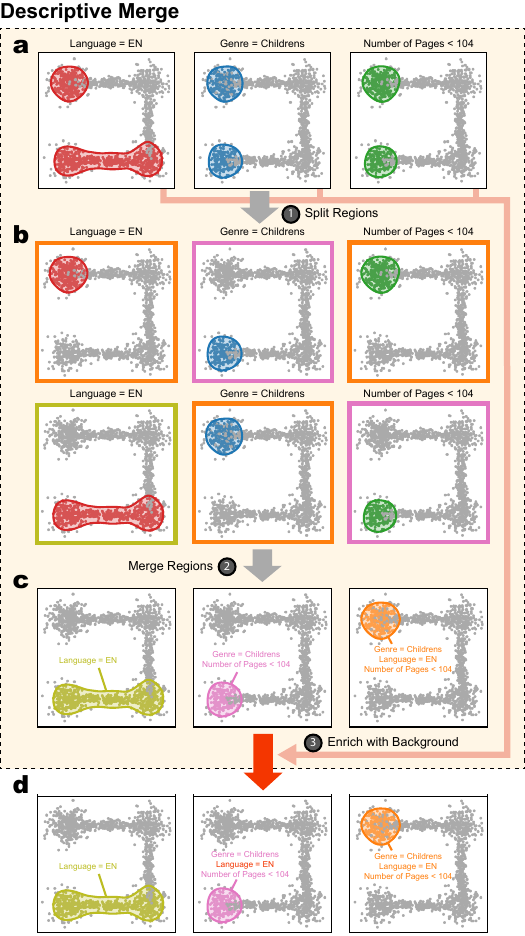}
  \caption{\label{fig:descriptive_merge}The descriptive merge, demonstrated on three region annotations (a). First, we split region annotations into disjoint regions (b). Then, we identify spatially overlapping region annotations and merge them into region annotation groups (c). Finally, each region annotation group is enriched with background annotations (d). In this instance, the middle panel depicting the pink region receives an additional descriptor, highlighted in red.}
\end{figure}

To identify candidate regions and their characteristic feature values, our method detects sets of region annotations with a high degree of overlap and merges them into annotation groups. Each resulting group then describes a particular region using the combination of features and feature values of each included region annotation.
As in contrastive explanations, we use the region annotations produced by the region construction procedure outlined in Section~\ref{sec:region-construction} as our starting point.

To group overlapping region annotations, we introduce the \textit{descriptive merge}, which consists of three main steps illustrated in Fig.~\ref{fig:descriptive_merge}. First, we split region annotations containing multiple disjoint regions into separate region annotations. For instance, the three region annotations shown in Fig.~\ref{fig:descriptive_merge}.a each contain two disjoint regions, which are split into separate annotations in Fig.~\ref{fig:descriptive_merge}.b.
  
Second, we merge sets of overlapping region annotations into region annotations groups.
Two region annotations can be merged if their regions $r_i$ and $r_j$ exhibit a sufficiently high degree of \textit{minimum overlap}:
\begin{equation}
\text{min\_overlap}(r_i, r_j) = \min \left \{ \frac{|r_i \cap r_j|}{|r_i|}, \frac{|r_j \cap r_i|}{|r_j|} \right \}. \notag
\end{equation}

Third, while the previous step merges region annotations exhibiting near-perfect overlap, these merged region annotations may not fully describe the characteristics of the samples within the region. For instance, the pink region in the middle panel of Fig.~\ref{fig:descriptive_merge}.c represents short children's books, obtained by merging the near-perfectly overlapping regions in Fig.~\ref{fig:descriptive_merge}.b highlighted in pink. However, these books are also fully contained within the yellow region corresponding to the English language in the left panel of Fig.~\ref{fig:descriptive_merge}.c. Hence, it is more informative to describe them as short, English children's books, as shown in Fig.~\ref{fig:descriptive_merge}.d, rather than simply as short children's books.
Here, English serves as an additional descriptor of the highlighted region rather than indicating where all English books appear in the embedding.
To incorporate these additional characteristics, the third step enriches the generated region annotations with rules from larger, sufficiently overlapping regions.
Importantly, we do not merge the regions themselves; instead, we augment smaller regions with rules from larger, overlapping ones.
We refer to this process as \textit{background enrichment}.

Step 2 resembles the post-hoc merging procedure described in Section~\ref{sec:post-hoc-merge}, in which the \textit{maximum overlap} had to exceed a predefined threshold. There, the maximum overlap ensured that smaller region annotations could be absorbed into larger, encompassing regions. Here, however, this approach would lead to misleading annotations.
Consider merging two region annotations $e_i$ and $e_j$ associated with rules $\varphi_i, \varphi_j$ and regions $r_i, r_j$ respectively.
If $r_j$ is a small region entirely contained within the larger region $r_i$, the points in $r_j$ will satisfy both rules $\varphi_i, \varphi_j$, while the remaining points in $r_i$ that are not in $r_j$ will not satisfy $\varphi_j$ but only $\varphi_i$.
However, the merged rule $\varphi_{\text{new}}$ would incorrectly indicate that all points within the merged region satisfy both $\varphi_i$ and $\varphi_j$, leading to a misleading annotation. To avoid this, we instead require a high degree of \textit{minimum overlap} between two regions, ensuring that both cover approximately the same data points and that the resulting annotation remains valid across all encompassed samples.

\subsubsection{Explanation Selection}
\label{sec:descriptive-explanations:selection}

\begin{figure*}[t!]
  \centering
  \includegraphics{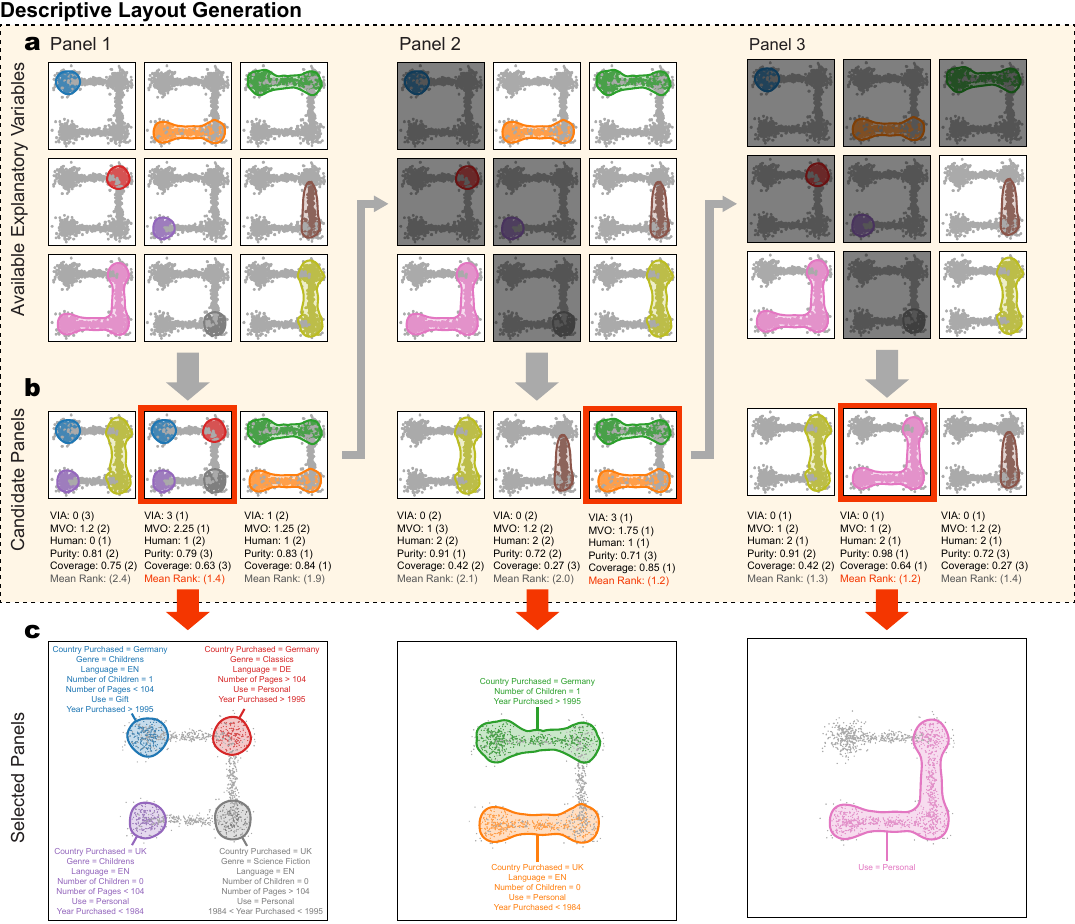}
  \caption{\label{fig:descriptive_generation}The descriptive layout generation follows an iterative approach: given an initial pool of region annotations in (a), we generate candidate panels by identifying their non-overlapping combinations. We show three such candidate panels in (b). Then, we score each panel along five different metrics, which are subsequently ranked and averaged. The panel with the highest mean rank, indicated in red, is added to the resulting layout, shown in (c). Finally, we remove the region annotations in the selected panel from the candidate pool. We then repeat the procedure until we obtain a specified number of panels or until the pool runs out of region annotations.}
\end{figure*}

Descriptive explanations are generated iteratively.
Given an initial pool of available region annotations, we identify all their potential combinations such that the overlap between them does not exceed some user-specified threshold.
Each candidate panel is scored across five criteria. The resulting scores are ranked and aggregated using a mean-rank strategy, similar to that in Section~\ref{sec:contrastive-selection}, and the highest-ranking panel is added to the final layout.
The region annotations from the selected panel are then removed from the candidate pool, and the procedure is repeated until we generate the requested number of panels $k$ or until the pool of available region annotations is exhausted. This procedure is illustrated in Fig.~\ref{fig:descriptive_generation}.

To efficiently identify combinations of non-overlapping region annotations, we formulate the task as an independent-set problem. We represent each region annotation as a node in a graph and place edges between them if their overlap exceeds a predetermined threshold. Then, the independent sets of this graph correspond to sets of non-overlapping region annotations, which are then considered candidate panels.

Descriptive explanations are ranked and selected based on five criteria.
The first three include the \textit{sample coverage}, i.e., what proportion of all data points are covered by the region annotations in the panel, the \textit{mean purity}, and the \textit{human attention score} from Section~\ref{sec:contrastive-selection}.
These metrics favor panels that offer broad coverage of the embedding space, contain reliable regions, and remain visually manageable, while softly preferring simpler panels over more crowded ones.
The fourth and fifth criteria ensure a degree of contrastivity, which, according to Miller~\cite{Miller2019}, should lead to more intuitive and satisfying explanations.
Although descriptive explanations may characterize regions using distinct sets of features, we prioritize panels where features recur across multiple region annotations. This enables users to compare feature values across regions within a single panel, without needing to reference other visual explanations.
To prioritize feature repetition, we score each panel according to its \textit{mean variable occurrence} (MVO), defined as the average number of times each feature appears across all region annotations within a particular panel.
Panels with higher MVO scores are preferred, as they facilitate within-panel comparisons.
We count the number of features that appear in all region annotations within a given panel using a dedicated \textit{variable in all} (VIA) score, favoring panels with higher VIA scores.
Together, these five metrics ensure the selection of high-quality region annotations that comprehensively cover the embedding space while aligning with human perceptual and cognitive preferences.

Across the full pipeline, from region construction to explanation generation, our approach involves several user-settable parameters. In practice, the results are robust across a range of parameter settings, and the default values generally yield high-quality explanations. Further details are provided in the parameter sensitivity study in the supplementary materials.

\section{Usage Scenarios}
\label{sec:case-studies}

To demonstrate the versatility and effectiveness of VERA, we present generated explanations across several diverse datasets. When a dataset includes a known class or target variable, we use it only as a post hoc variable for interpreting the embedding and do not use it to construct the embedding itself.

\subsection{IBM Employee Attrition Dataset}

The IBM Employee Attrition dataset~\cite{IbmEmployeeAttrition} contains records of 1,470 fictional employees, each characterized by 33 features. We obtain a two-dimensional embedding with the t-SNE algorithm implemented in the \textsf{openTSNE} library~\cite{Policar2024}.

\begin{figure*}[t]
  \centering
  \includegraphics{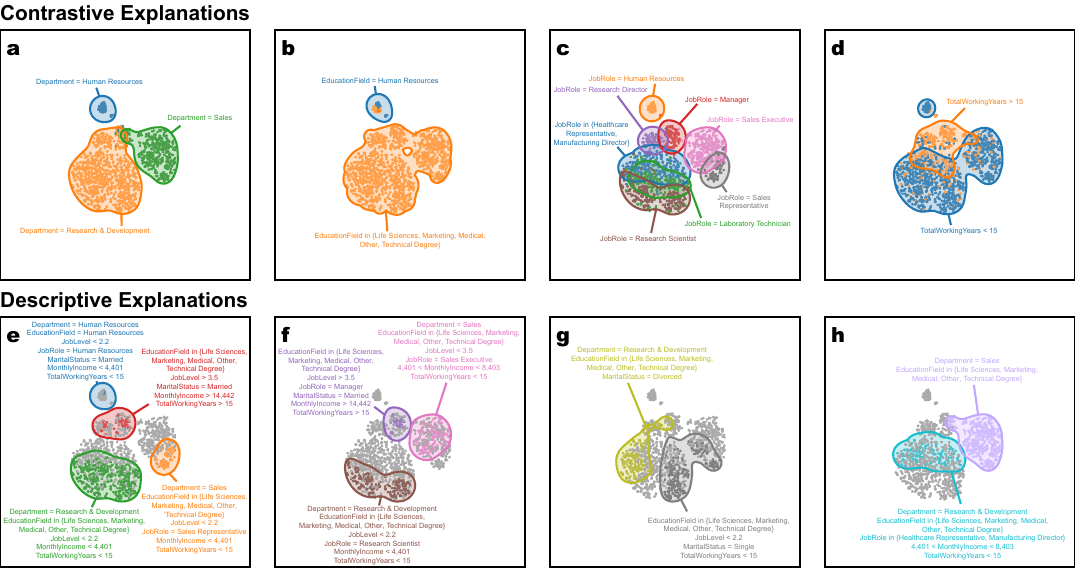}
  \caption{\label{fig:attrition}Explanations of the IBM Employee Attrition dataset. Panels (a-d) show four contrastive explanations corresponding to the spatially most correlated features with this particular embedding. The descriptive explanations shown in panels (e-h) characterize different regions of the embedding space.}
\end{figure*}

Fig.~\ref{fig:attrition} shows the four highest-ranked contrastive and descriptive explanations selected by VERA.
The first two contrastive explanations in Figs.~\ref{fig:attrition}.a and b reveal that the three clusters are primarily distinguished by employees' departments and education fields.
Fig.~\ref{fig:attrition}.c shows that the three clusters can be further subdivided into specific job roles, while Fig.~\ref{fig:attrition}.d shows that the number of total working years also corresponds to the samples' spatial positions.
Together, these four panels already offer a meaningful interpretation of the three main clusters uncovered by the t-SNE algorithm.

The descriptive explanations depicted in Figs.~\ref{fig:attrition}.e-h offer descriptions of specific regions of the embedding space. For instance, the blue cluster in Fig.~\ref{fig:attrition}.e lists the characteristics of the human resources department and primarily includes junior, married employees with lower salaries. In contrast, the red cluster contains more senior, married individuals with higher salaries. Cross-referencing Fig.~\ref{fig:attrition}.c reveals that most employees in the red cluster hold managerial or research director roles. We can similarly inspect the other regions to gain further insight into the structure captured by the embedding.

Notice that many of the identified regions in Fig.~\ref{fig:attrition} do not correspond to discrete clusters but rather represent regions within or across clusters. For instance, the gray region in Fig.~\ref{fig:attrition}g corresponds to single employees with less than 15 years of experience and low job levels—likely younger, junior staff. Conversely, the yellow region primarily includes divorced employees, for whom job level and total working years are unspecified. This demonstrates VERA’s ability to characterize not only entire clusters but also informative subregions within them.

\subsection{Abalone Dataset}
\label{sec:case-studies:abalone}

\begin{figure}[htb]
  \centering
    \includegraphics{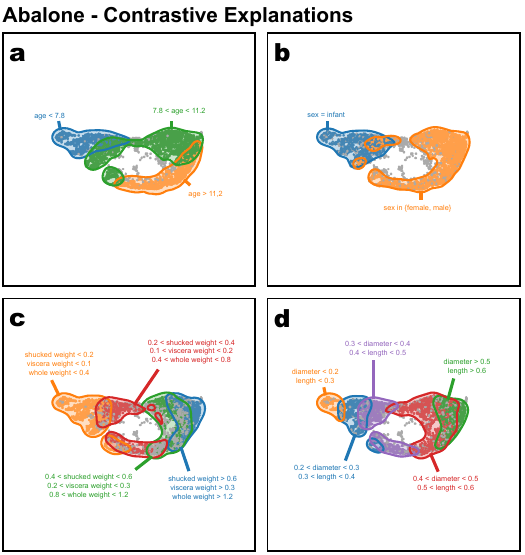}
    \caption{\label{fig:abalone}Contrastive explanations for a UMAP embedding of the Abalone dataset. Panels (a-d) highlight selected features, showing the spatial distribution of age, sex, and physical measurements, as well as correlations between features related to animal weight and size.}
\end{figure}

The Abalone dataset~\cite{Nash1994} contains 4,177 samples of abalones -- marine mollusks with distinctive ear-shaped shells. Each sample is characterized by eight features, including physical measurements of the body and shell and biological attributes such as age and sex. We generate a two-dimensional embedding with the UMAP algorithm using the \textsf{umap-learn} library~\cite{McInnes2018}. The embedding reveals an elongated structure extending from left to right.

Fig.~\ref{fig:abalone} shows four selected contrastive explanations generated by VERA.
The explanations reveal that the elongated shape reflects the size and age of the abalones: younger abalones are smaller and lighter, while older abalones are larger and heavier. Fig.~\ref{fig:abalone}.b also suggests that the sex of the animals does not affect their size.
Figs.~\ref{fig:abalone}.c-d highlight the benefits of the contrastive merge procedure. Fig.~\ref{fig:abalone}.c combines shucked weight, viscera weight, and whole weight into a single explanation, while Fig.~\ref{fig:abalone}.d combines shell diameter and length.
This not only reduces the number of explanations from five to two but also reveals the strong correlation between the three merged features.

\subsection{Iris and Penguin Datasets}

\begin{figure}[ht]
  \centering
  \includegraphics{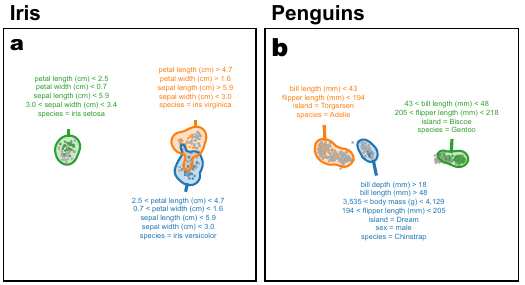}
  \caption{\label{fig:iris-penguins}Descriptive explanations of the Iris (a) and Penguin (b) datasets. Each explanation provides a comprehensive, high-level overview of the clusters revealed in the embedding.}
\end{figure}

The Iris~\cite{Anderson1936} and Penguin~\cite{Horst2020} datasets are widely used benchmarks in data visualization and machine learning due to their clear clustering structure and overall simplicity.
Fig.~\ref{fig:iris-penguins} shows one descriptive explanation for each dataset, providing a concise overview of the main patterns and relationships.
In the Iris dataset (Fig.~\ref{fig:iris-penguins}.a), VERA identifies regions corresponding to the three species of iris flowers, each characterized by distinct combinations of sepal and petal measurements. For the Penguin dataset (Fig.~\ref{fig:iris-penguins}.b), VERA identifies clusters representing the three penguin species, separated by features such as flipper length and body mass.
For simple datasets such as these containing relatively few features and having relatively well-defined clusters, a single descriptive explanation is often sufficient to convey the entire embedding structure.

\section{User Evaluation}
\label{sec:user-study}

To evaluate the effectiveness of our approach, we conducted a user study comparing participants' speed and accuracy in reasoning about low-dimensional embeddings using an interactive exploration tool versus our static VERA explanations.

\subsection{Study Design}
\label{sec:user-study:design}

We selected the Orange data mining toolkit~\cite{Demsar2013} as our interactive baseline because of its ease of use and support for exploratory analysis through interactive brushing and linked views.
Although not explicitly designed for interpreting embeddings, Orange includes standard visualization tools for exploring two-dimensional embeddings. Many related approaches based on feature distribution plots~\cite{Stahnke2016,Eckelt2022} and discriminative models~\cite{Bibal2021,Marcilio2021} can be easily reproduced in Orange using only a handful of widgets. We provide several example workflows in the supplementary materials.

We prepared two groups of three real-world datasets each, selected to vary in complexity.
The first dataset group $D_1$ comprises the Iris~\cite{Anderson1936}, Employee Attrition~\cite{IbmEmployeeAttrition}, and Raisin~\cite{RaisinDataset} datasets. The second dataset group $D_2$ includes the Penguins~\cite{Horst2020}, Titanic~\cite{Dawson1995}, and Wine Quality~\cite{Cortez2009} datasets. For each dataset, participants answered a series of typical exploratory questions about two-dimensional embeddings, such as: ``What makes this cluster different from the remaining clusters?'', ``What characterizes the samples in this cluster?'', and ``Do you notice any trend along this region?''.

We recruited 57 bachelor's students enrolled in the Introduction to Machine Learning course at the Faculty of Computer and Information Science at the University of Ljubljana.
Participation was purely voluntary, and no personally identifiable information was collected.
Prior to the study, students were already familiar with Orange but not with VERA explanations, so we provided a brief tutorial on both tools before beginning the experiment. This prior familiarity with Orange reduced additional interface-learning overhead for the interactive baseline.
Participants were randomly assigned to one of two experimental groups. The first group $G_1$ first used Orange to answer questions related to datasets $D_1$, then used VERA explanations for questions relating to datasets in $D_2$. Conversely, group $G_2$ first used VERA on $D_1$ and then Orange on $D_2$. On average, students answered $29.2$ of the $32$ questions ($\text{SD}= 4.3$, $\text{SE}=0.56$), amounting to a total of $1,666$ responses.

To assess response time and accuracy, we measured how long each participant took to answer each question and whether the response was correct. After completing both dataset groups, participants filled out a final questionnaire to provide qualitative feedback on both tools. The complete catalog of study questions, participant-facing visual materials, and questionnaires are provided in the supplementary materials.

\subsection{Results}
\label{sec:user-study:results}

The results show that participants using VERA explanations performed as accurately as those using Orange but required significantly less time.
The difference in user performance was not statistically significant ($t$-test $p=0.084$), with 91\% ($\text{SE}=0.9\%$) of participants answering correctly when using Orange and 93\% ($\text{SE}=1.0\%$) when using VERA.
However, participants in both user groups answered significantly faster when using VERA explanations ($p<0.005$). On average, participants using Orange took $29$ minutes ($\text{SD}= 11.5$, $\text{SE}=2.1$) to complete questions related to $D_1$ and $23$ minutes ($\text{SD}= 6.0$, $\text{SE}=1.1$) to complete questions related to $D_2$. Conversely, participants using VERA explanations required $21$ minutes ($\text{SD}= 6.3$, $\text{SE}=1.2$) and $15$ minutes ($\text{SD}=5.4$, $\text{SE}=1.1$) for each dataset group, representing a consistent increase in speed of approximately 33\%.
These results confirm that participants in both groups were able to answer the questions with roughly the same degree of accuracy but at a much faster rate when using VERA explanations.
We consider this gain practically meaningful, since analysts typically inspect many embeddings and perform similar interpretation tasks; in that setting, a 33\% reduction in response time substantially reduces cumulative effort.

Interestingly, when asked to list characteristic or distinguishing features for particular regions, participants often mentioned more features when using VERA explanations ($p=0.04$). When generating explanations using Orange, participants mentioned an average of $2.0$ ($\text{SD}= 1.29$, $\text{SE}=0.07$) features in their answers, while this increased to $2.2$ ($\text{SD}= 1.33$, $\text{SE}=0.07$) when using VERA explanations. This difference may stem from several factors. When using an interactive tool, users must manually inspect multiple features to assess their relationship to the embedding space. Participants often stopped their exploration after identifying one or two informative features, potentially overlooking additional relevant ones. VERA explanations, on the other hand, provide all relevant explanations at once. Although these explanations display only a limited number of features, VERA’s ranking mechanism ensures that the most salient ones appear first. Thus, the user simply \textit{sees} all the features and cannot miss them by accident.

In the final feedback questionnaire, we asked participants to document their experience with both tools. Consistent with our quantitative results, most participants (91\%, $\text{SE}=4\%$) reported that they were able to answer questions more quickly using the VERA explanations. The majority of participants (77\%, $\text{SE}=6\%$) also confirmed that the static explanations were as informative as those produced with Orange.
Surprisingly, when asked which tools they trusted more, 50\% ($\text{SE}=8\%$) of participants said they trusted both tools equally, 20\% ($\text{SE}=6\%$) said they trusted VERA explanations more, and 30\% said they trusted Orange more ($\text{SE}=7\%$). 
We anticipated that users would trust full-featured, interactive toolkits like Orange more than static visual explanations, as interactive tools enable multiple validation mechanisms and allow users to drill down into specific visualization components. VERA explanations, on the other hand, are static and offer no such options, forcing the user to simply trust the resulting visualizations.

Finally, participants rated the usefulness of both types of VERA explanations on a five-point Likert scale and indicated which type they found more informative. Consistent with Miller's~\cite{Miller2019} findings, most participants (75\%, $\text{SE}=6.6\%$) selected contrastive explanations as the most informative, while 15\% ($\text{SE}=5.4\%$) found both types of explanations equally informative, and 10\% ($\text{SE}=4.7\%$) found descriptive explanations more informative.
Contrastive explanations received a higher overall rating of $4.3$ ($\text{SD}= 0.89$, $\text{SE}=0.13$), while descriptive explanations received a lower overall rating of $3.2$ ($\text{SD}= 1.06$, $\text{SE}=0.16$).
This result does not suggest that descriptive explanations lack utility. Rather, it indicates that users generally perceive contrastive explanations as more informative.

\section{Scalability and Limitations}
\label{sec:scalability}

We assess scalability from two perspectives, algorithmic complexity and visual limitations.

\begin{figure}[t]
  \centering
  \includegraphics{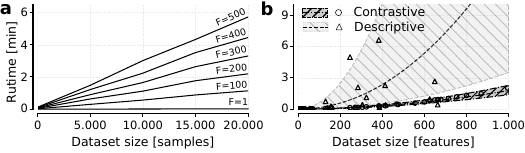}
  \caption{\label{fig:benchmarks} Single-core performance of VERA on a consumer-grade M1 MacBook Pro on different-sized synthetic datasets. (a) VERA scales linearly in the number of samples $N$, as shown by the six near-straight lines, corresponding to datasets with a different number of features $F$. (b) Runtime with respect to the number of region annotations (after post-hoc merging and filtering) $F$ for contrastive and descriptive explanation generation.}
\end{figure}

\textit{Algorithmic complexity} can be analyzed with respect to the number of samples $N$ and features $F$.
In VERA, the samples are used only during the region construction step described in Section~\ref{sec:region-construction}, where a kernel density estimate is computed for each feature. The KDE can be efficiently computed in linear time using an FFT-accelerated interpolation-based approach, yielding a linear time complexity $\mathcal{O}(N)$.
Since the remaining steps of VERA operate solely on the extracted region annotations, the overall time complexity with respect to $N$ is linear, reflected in Fig.~\ref{fig:benchmarks}.a. Accurate KDE estimates can also be obtained through dataset subsampling, further improving scalability.

The number of features $F$ has a greater impact on runtime.
After extracting initial region annotations, VERA performs a post-hoc merging step for each feature, requiring linear time $\mathcal{O}(F)$.
Both contrastive and descriptive explanations construct an overlap graph of region annotations by computing pairwise overlaps, an operation with quadratic time complexity $\mathcal{O}(F^2)$. The overlap graph is then used for both contrastive and descriptive merge steps.
In contrastive explanations, the remaining step is a linear time panel ranking, resulting in an overall time complexity of $\mathcal{O}(F^2)$, as can be seen in Fig.~\ref{fig:benchmarks}.b.
In descriptive explanations, we use the merged region annotations to construct a second overlap graph and use its independent sets as candidate panels. Finding independent sets is NP-hard, requiring $\mathcal{O}(2^F)$ time.
In practice, runtime is typically much lower and varies across datasets, as shown in Fig.~\ref{fig:benchmarks}.b.
This reduction stems from the descriptive merge step, which often substantially decreases the number of candidate region annotations, yielding smaller graphs and faster independent set computations.
For datasets with many features, a greedy graph-coloring heuristic yields good approximate solutions in linear time. In this case, graph construction ($\mathcal{O}(F^2)$) dominates the descriptive explanation generation time.
Although both explanation types exhibit superlinear complexity, VERA remains practical for typical real-world datasets, completing in seconds to a few minutes on standard hardware. 

\textit{Visual limitations} primarily affect descriptive explanations, especially when the feature count is large. In these cases, VERA may produce long lists of region descriptors -- particularly when background enrichment is used -- which pose two key challenges.
First, long descriptor lists impose a cognitive burden on users, who must manually inspect and interpret the output. Second, these lists consume screen space, leading to visual clutter: labels may overlap with each other or with data points, extend off-screen, or otherwise obstruct the view.
In our current formulation, VERA is best suited for datasets with up to several hundred features, where meaningful explanations can be expressed in terms of a relatively small subset of variables. For datasets with thousands or tens of thousands of features, directly applying VERA becomes less effective, both computationally and visually, as the number of candidate region annotations grows and the resulting explanations become difficult to interpret.
Future work could address these issues by leveraging large language models (LLMs) to automatically generate concise, informative summaries of lengthy region descriptors.

\section{Conclusion}

With the growing popularity of two-dimensional embeddings in data visualization and their widespread use in exploratory data analysis and scientific communication, there is a growing need for complementary tools to interpret, validate, and compare these embeddings.
We introduced \textit{Visual Explanations via Region Annotation (VERA)}, an automated method that enables users to rapidly interpret two-dimensional embeddings through static visual explanations.
VERA draws on insights from the social sciences to design visual explanations that align with human cognitive processes, producing outputs that are both informative and intuitively satisfying.
VERA addresses several limitations of existing interactive and static methods, combining the efficiency of static visualization with the interpretive richness of interactive exploration in common exploratory data analysis tasks.
Despite its effectiveness, VERA opens several promising directions for future research.
Extending VERA to handle diverse data modalities, including image, video, text, and graph data, improving scalability to very large feature sets through feature aggregation or abstraction, and exploring integration with interactive analytics platforms all represent promising directions. Furthermore, recent advances in LLMs for automatic summarization could be leveraged to generate more concise and relevant annotations.
Overall, VERA represents a significant step toward fast, effective, and cognitively aligned explanations of two-dimensional nonlinear embeddings.

\section*{Code and Supplementary Material Availability}

Our open-source Python implementation of VERA is freely available at \url{https://github.com/pavlin-policar/vera} and is licensed under the BSD 3-clause license. The replication and supplementary materials are available at \url{https://github.com/pavlin-policar/vera-paper}.

\section*{Acknowledgments}
This work was supported by the Slovenian Research Agency grants P2-0209 and V2-2272.
 
\bibliographystyle{ieeetr}
\bibliography{main}


 





\end{document}